%% file: main.tex
\documentclass[letterpaper, 10 pt, conference]{ieeeconf}  % Comment this line out if you need a4paper

\IEEEoverridecommandlockouts                              % This command is only needed if 
\usepackage{microtype}
\usepackage{graphicx}
\usepackage{subcaption}
\usepackage{booktabs} % for professional tables
\usepackage{multirow}
\usepackage{hyperref}
\usepackage{algorithm}
\usepackage{algorithmic}

\usepackage{amsmath}
\usepackage{amssymb}
\usepackage{mathtools}
\usepackage{amsthm}
\usepackage{wrapfig}
\usepackage{colortbl}
\usepackage{authblk}
\usepackage[dvipsnames,table]{xcolor}
\usepackage{arydshln}
\definecolor{lightgray}{gray}{0.8}
\definecolor{lightorange}{rgb}{1.0, 0.917, 0.655}
\definecolor{lightblue}{rgb}{0.9,0.9,1.0}

\colorlet{tlightorange}{lightorange!10} % 50% light orange
\colorlet{tlightblue}{lightblue!10} % 50% light blue
\definecolor{HighlightBest}{RGB}{175, 255, 255}
\definecolor{HighlightSecond}{RGB}{255, 234, 167}
\definecolor{HighlightWorst}{RGB}{255, 190, 118}

\usepackage[capitalize,noabbrev]{cleveref}

\newcommand{\nopar}[1]{%
  \par % Ensure the previous paragraph is closed
  \noindent % Start the next paragraph without indentation
  #1 % Insert the content
  \par % End the paragraph
}

\usepackage[textsize=tiny]{todonotes}
\usepackage{bm}

 \newcommand{\spm}[2][normal]{%
     \ifthenelse{\equal{#1}{bold}}%
     {\ensuremath{{\scriptstyle\boldsymbol{\pm} \mathbf{#2}}}}%
     {\ensuremath{{\scriptstyle \pm #2}}}%
 }

 \newcommand{\entry}[3][normal]{%
     \ifthenelse{\equal{#1}{bold}}%
     {\textbf{#2} \ensuremath{{\scriptstyle\boldsymbol{\pm} \mathbf{#3}}}}%
     {#2 \ensuremath{{\scriptstyle \pm #3}}}%
 }

\title{\LARGE \bf
FLoRA: Sample-Efficient Preference-based RL via Low-Rank \\ Style Adaptation of Reward Functions
}

\author{Daniel Marta$^{*,\dagger}$, Simon Holk$^{*,\dagger}$, Miguel Vasco$^{\dagger}$, Jens Lundell$^{\dagger}$, Timon Homberger$^{\dagger}$, Finn Busch$^{\dagger}$,\\ Olov Andersson$^{\dagger}$, Danica Kragic$^{\dagger}$ and Iolanda Leite$^{\dagger}$% <-this % stops a space
\thanks{$^*$Shared first-authorship.}% <-this % stops a space
\thanks{$^{\dagger}$KTH Royal Institute of Technology, Sweden,
        \{dlmarta, sholk, miguelsv, jelundel,timonh,flbusch,olovand,dani,iolanda\}@kth.se}}%

\begin{document}

\maketitle
\thispagestyle{empty}
\pagestyle{empty}

\input{settings}
%%%%%%%%%%%%%%%%%%%%%%%%%%%%%%%%%%%%%%%%%%%%%%%%%%%%%%%%%%%%%%%%%%%%%%%%%%%%%%%%
\begin{abstract}
    Preference-based reinforcement learning (PbRL) is a suitable approach for \emph{style} adaptation of pre-trained robotic behavior: adapting the robot's policy to follow human user preferences while still being able to perform the original task. However, collecting preferences for the adaptation process in robotics is often challenging and time-consuming. In this work we explore the adaptation of pre-trained robots in the low-preference-data regime. We show that, in this regime, recent adaptation approaches suffer from \emph{catastrophic reward forgetting} (CRF), where the updated reward model overfits to the new preferences, leading the agent to become unable to perform the original task. To mitigate CRF, we propose to enhance the original reward model with a small number of parameters (low-rank matrices) responsible for modeling the preference adaptation. Our evaluation shows that our method can efficiently and effectively adjust robotic behavior to human preferences across simulation benchmark tasks and multiple real-world robotic tasks. We provide videos of our results and source code at \url{https://sites.google.com/view/preflora/}.
\end{abstract}

%%%%%%%%%%%%%%%%%%%%%%%%%%%%%%%%%%%%%%%%%%%%%%%%%%%%%%%%%%%%%%%%%%%%%%%%%%%%%%%%
\section{Introduction}
\label{sec:intro}

\begin{figure*}[t]
	\centering
	\includegraphics[width=\linewidth]{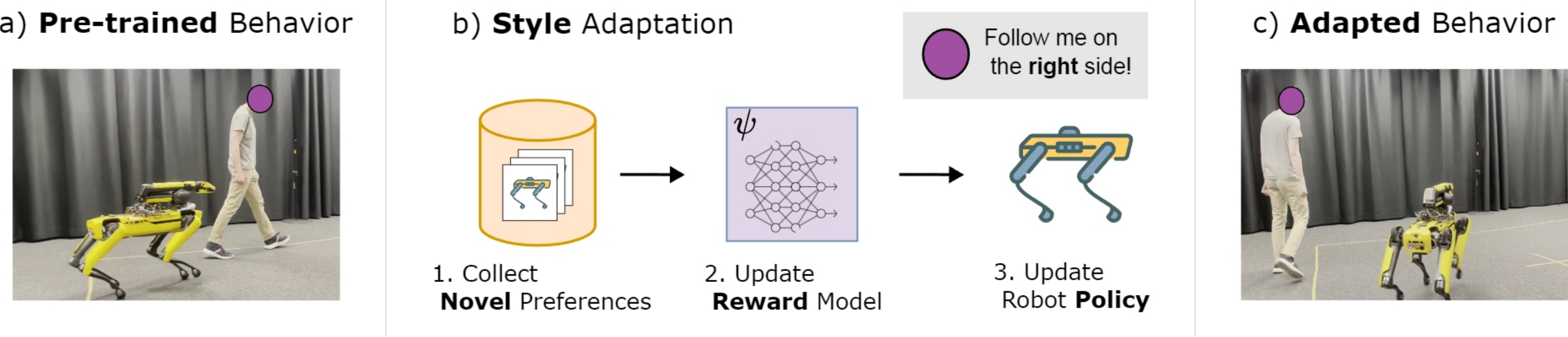}
\caption{\textbf{Style Adaptation of Robotic Behavior.} We focus on the \emph{style adaptation} of pre-trained robot behavior, where the goal of the adaptation is to adjust the behavior of the agent accordingly to human preferences, while still being able to perform the original task: (a) a four-legged mobile robot is pre-trained to follow behind a human user; (b) we collect a small number of novel human preferences (e.g., following the user on his right) to train a reward model and the adapted policy of the robot, while \emph{maintaining the ability} to follow the human and avoiding collisions; c) we execute the style-adapted policy following the human preferences.}
\vspace{-10pt}
\label{fig:intro}
\end{figure*}

Reinforcement learning has become a general approach to train agents due to the emergence of large-scale datasets of interaction data~\cite{padalkar2023open}, and realistic simulation platforms~\cite{liang2018gpu, mittal2023orbit}. However, adapting pre-trained general behavior to human preferences remains a challenge, mainly due to the difficulty of handcrafting suitable reward functions. Preference-based reinforcement learning (PbRL)~\cite{wirth2017survey} provides a solution to this problem by guiding the adaptation process through pairwise comparisons of human preferences over the behavior of the agent~\cite{christiano2017deep,ibarz2018reward,ziegler2019fine}. In this work, we explore PbRL for \emph{style} adaptation\footnote{Our notion of style adaptation is akin to adapting generative models of text~\cite{li2018measuring,devlin2018bert,ziegler2019fine,hu2021lora,shah2023ziplora} and images~\cite{karras2019style,zhong2024multi} to human preferences.}, where the goal of the adaptation process is to adjust the behavior of the agent (e.g. through adjusting its reward model) accordingly to human preferences, while still being able to perform the original task.
In Figure~\ref{fig:intro} we show the style adaptation of a four-legged robot, whose pre-trained behavior is to follow the human user and we adapt its behavior to do so always on the right side of the user.

We focus on the style adaptation of pre-trained \emph{robotic} reward models to human preferences. Behavior adaptation in robotics raises a distinctive problem: collecting human preferences is often challenging and time-consuming. This restricts the style adaptation process to rely only on small datasets of preferences, often consisting of a few hundred to a few thousand behavior comparisons~\cite{christiano2017deep,lee2021pebble,hejna2023few}. We show that, in this small-preference-data regime, recent proposed methods for adaptation result in \emph{catastrophic reward forgetting} (CRF), where the adaptation of a reward model to a new preference distribution results in severe degradation of the original reward distribution. This degradation leads to sub-optimal robot behavior or, in extreme cases, to the inability of the robot to perform the original task~\cite{dalal2023plan,dai2023empowerment}.

We propose to address CRF through the lens of \emph{intrinsic dimensionality}~\cite{aghajanyan2021intrinsic}, that suggests that only a small number of parameters are needed to fine-tune pre-trained reward models to human preferences. As we highlight in Section~\ref{sec:challenge}, in the low-data regime, human preferences over robotic behavior often lie in a small region of the entire state-action space of the agent. Motivated by these results, we propose to use low-rank adaptation~\cite{hu2021lora} to learn a compact representation of the reward function over the distribution of the provided preferences, instead of fine-tuning the pretrained reward model.  The learned representation conditions the style adaptation process without changing the weights of the original reward model, thus mitigating CRF. We call our approach Pre\textbf{F}erence-based Reinforcement Learning via \textbf{Lo}w-\textbf{R}ank Style \textbf{A}daptation, \textbf{FLoRA}. FLoRA is agnostic to the underlying PbRL algorithm and can be naively employed alongside other standard techniques, e.g. semi-supervised learning, for the style adaptation process.

% \begin{figure*}[t]
% 	\centering
% 	\includegraphics[width=\linewidth]{figures/introduction/intro_icra_real.pdf}
% \caption{\textbf{Style Adaptation of Robotic Behavior.} We focus on the \emph{style adaptation} of pre-trained robot behavior, where the goal of the adaptation is to adjust the behavior of the agent accordingly to human preferences, while still being able to perform the original task: (a) a four-legged mobile robot is pre-trained to follow behind a human user; (b) we collect a small number of novel human preferences (e.g., following the user on his right) to train a reward model and the adapted policy of the robot, while \emph{maintaining the ability} to follow the human and avoiding collisions; c) we execute the style-adapted policy following the human preferences.}
% \vspace{-10pt}
% \label{fig:intro}
% \end{figure*}

We evaluate the style adaptation performance and sample efficiency of FLoRA across standard control benchmarks in simulation and on two real-world robotic platforms. The results show that our method allows reward functions (and, thus, policies) to efficiently adjust to human preferences without suffering from CRF, often requiring \emph{less than 100 human preferences}. We also evaluate the adaptation performance of FLoRA in real-world tasks performed by a 7-DoF manipulator and a four-legged mobile robot.

To the best of our knowledge, this is the first work to explore the use of low-rank adaptation for adapting pre-trained robot behavior. In summary, our contributions are:
\begin{itemize}
\item \textbf{CRF in Style Adaptation.} We show that employing fine-tuning in a low-sample regime with preference-based RL leads to a decline in overall policy quality and a loss of previously acquired baseline behavior;
\item \textbf{FLoRA.} A novel method that enables a sample-efficient, modular approach to adapting the style of pretrained robotic behavior, agnostic to the underlying PbRL algorithm. Our approach learns a low-rank representation of the novel preference reward distribution, maintaining the weights of the original reward model.
\item \textbf{Adaptability and Sample Efficiency.} Our evaluation reveals that FLoRA can effectively adjust to human preferences in the low-data regime in various control tasks, both in simulation and on two real-world robotic platforms, while mitigating CRF in comparison with other fine-tuning techniques.
\end{itemize}

\section{Related Work}
\label{sec:related}

\nopar{\textbf{Human preferences in robot learning} \quad Preference-based RL~\cite{wirth2017survey} has become a suitable approach for formulating complex objectives into reward functions. In robot learning, preferences have been used in multi-task settings~\cite{hejna2022few}, collaborative tasks~\cite{zhao2023learning}, and alongside language~\cite{holk2024predilect}. Nonetheless, the need for large amounts of human feedback in these applications has revealed constraints in their feasibility for real-world robotics and other sophisticated environments~\cite{liang2022reward,park2022surf, hu2022explaining,hejna2023few}. Current trends in preference learning usually place few limitations on the type of reward functions, often resulting in a process akin to iterative inverse reinforcement learning (IRL)~\cite{amin2017repeated}. These reward functions might be built from the ground up, following a 'blank slate' (\textit{tabula rasa}) methodology~\cite{christiano2017deep}, or be initially bootstrapped via imitation learning~\cite{ibarz2018reward}. Current approaches to address sample efficiency rely on re-utilizing the entire collection of preferences~\cite{lee2021pebble,kim2023preference}, allowing the reward function to adapt based on the performance of the Q-function~\cite{liu2022meta}, training on a multitude of tasks to fine-tune on a similar task~\cite{hejna2023few}, or leveraging the inherent transitivity of human preferences~\cite{hwang2024sequential}. We complement the aforementioned works by enabling the reuse of the original reward model through low-rank style adaptation, providing a modular approach that is completely agnostic to the underlying PbRL and RL algorithms used.
}

\nopar{\textbf{Adaptation of reward functions} \quad The complexity of modeling reward functions with human feedback is emphasized by \cite{tien2022study}, who, through comprehensive empirical studies, demonstrated how the introduction of irrelevant features and increased model complexity can lead to misunderstandings about the actual reward function, even when thousands of pairwise preferences are used. This situation might create spurious correlations, potentially leading to the manipulation of rewards~\cite{amodei2016concrete,hadfield2017inverse} or causing a shift in the distribution~\cite{de2019causal}. Multiple reward functions can fit preferences within a dataset, suggesting that reward functions are only partially identifiable~\cite{skalse2023invariance}. Moreover, \cite{mckinney2023fragility} highlight the vulnerability of learned reward models to the design and query composition, which can hinder their ability to train new agents from scratch in diverse settings effectively. Given the nature of low sample sizes usually employed in preference-based RL, FLoRA contributes a flexible approach for adapting reward functions that captures the distribution of the new query distribution without impacting general task performance.
}
% \cite{kannan2023deft} fine-tunningggg

\nopar{\textbf{Overoptimization of reward models} \quad The issue of policy and reward model overfitting has been a key focus in RL literature~\cite{zhang2018dissection, cobbe2019quantifying, song2019observational, kannan2023deft}, with some works being adamant about reducing overoptimization of reward functions~\cite{dai2023empowerment, dalal2023plan, gleave2022uncertainty}. Particularly in preference-based RL, overfitting a reward function to a subset of the state space can result in significant drawbacks, such as catastrophic forgetting and diminished policy performance. In the context of RL from human feedback (RLHF), the process of fine-tuning RL models has been shown to reduce the variety of outputs generated by these models~\cite{glaese2022improving, go2023aligning, gao2023scaling}, an effect known as \textit{mode collapse}. The observed reduction in output diversity during RL fine-tuning is attributed to the propensity for favoring sequences that yield higher scores with increased probability rather than maintaining probabilities that are consistent with the training distribution~\cite{casper2023open}. In this work, we address overfitting by constraining gradient updates to fewer parameters and improve generalization by retaining the primary reward model.
}

% \begin{figure*}[t]
% 	\centering
% 	\includegraphics[width=\linewidth]{figures/crf/icra_crf.png}
% \caption{\textbf{The Problem of CRF.} t-SNE projection of state-action pairs sampled from the ~\texttt{Drawer-Close} simulation environment~\cite{yu2020meta} with their associated reward values. We plot the normalized reward value (higher is preferred), from green (higher) to red (lower), predicted for each state-action pair by the true reward function (orange) and the different style adapted reward models (purple).}
% 	\label{fig:visual}
% \end{figure*}

\begin{figure*}[t]
    \centering
    \begin{subfigure}[t]{0.28\textwidth}
        \centering
        \includegraphics[height=3cm]{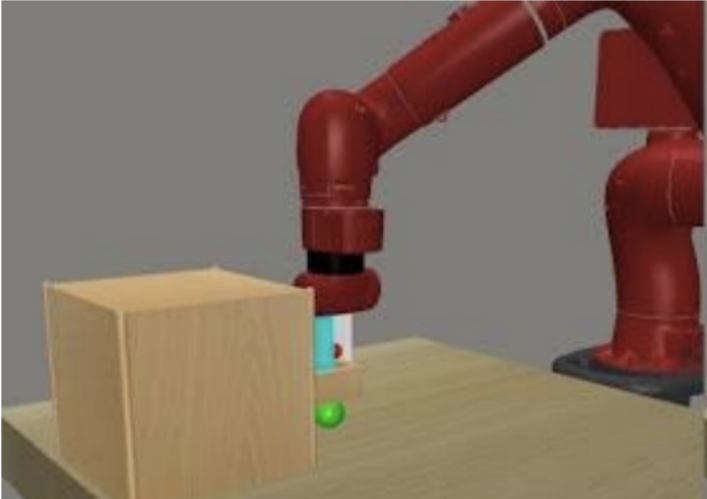}
        \caption{\texttt{Drawer-Close}~\cite{yu2020meta}}
    \end{subfigure}%
    \hfill
    \begin{subfigure}[t]{0.15\textwidth}
        \centering
        \includegraphics[height=3cm]{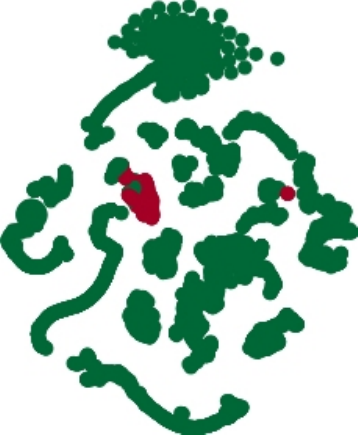}
        \caption{Original Reward}
    \end{subfigure}%
    \hfill
    \begin{subfigure}[t]{0.15\textwidth}
        \centering
        \includegraphics[height=3cm]{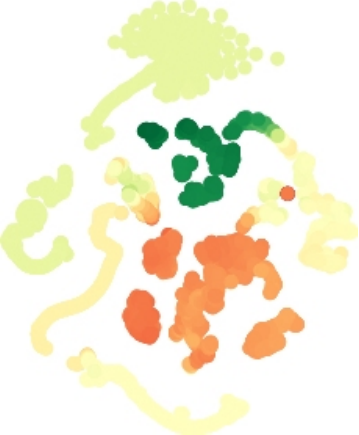}
        \caption{Fine-tuning~\cite{hejna2023few}}
    \end{subfigure}%
    \hfill
    \begin{subfigure}[t]{0.15\textwidth}
        \centering
        \includegraphics[height=3cm]{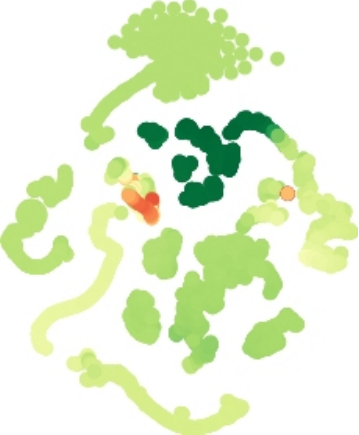}
        \caption{SURF~\cite{park2022surf}}
    \end{subfigure}%
    \hfill
    \begin{subfigure}[t]{0.15\textwidth}
        \centering
        \includegraphics[height=3cm]{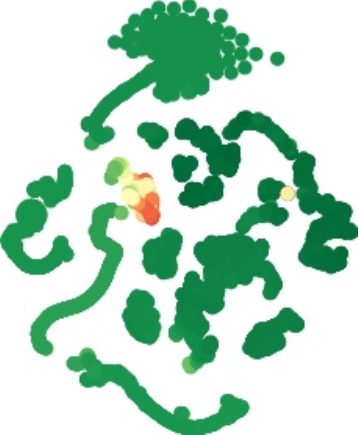}
        \caption{\textbf{FLoRA} (Ours)}
    \end{subfigure}
    \caption{\textbf{The Problem of CRF.} t-SNE projections of state-action pairs sampled from the \texttt{Drawer-Close} simulation environment (a) with associated reward values. We plot the normalized reward value (higher is preferred), from green (higher) to red (lower), predicted for each state-action pair by the true reward function (b) and by different adapted reward models (c-e). FLoRA uniquely adapts the reward model to new human preferences while preventing catastrophic forgetting.}
    \label{fig:visual}
    \vspace{-15pt}
\end{figure*}

\section{Background}
\label{sec:background}

%\textbf{Reinforcement Learning} \quad  Learning to perform decision-making tasks through trial-and-error interaction between an agent and its environment can be abstracted as a reinforcement learning (RL) problem ~\cite{sutton2018reinforcement}. The standard formalization of a RL problem is through a a Markov decision process (MDP) $\mathcal{M}$, instantiated as a tuple $\mathcal{M} = (\mathcal{S}, \mathcal{A}, P, r, \gamma)$, where $\mathcal{S}$ is the state space of the agent; $\mathcal{A}$ is the action space of the agent; the probability transition function $P(s_{t+1} \mid s_t, a_t)$ models the change of the environment when the agent takes an action $a_t \in \mathcal{A}$ in state $s_t \in \mathcal{S}$; the reward function $r(s_t, a_t)$ defines the immediate reward collected by the agent when performing action $a_t$ in state $s_t$; and the discount factor $\gamma \in [0, 1]$ weighs the relative importance of present and future rewards for the agent's learning process. Through trial and error, the agent aims to learn a policy $\pi : \mathcal{S} \times \mathcal{A} \to [0, 1]$, where $\pi(a_t \mid s_t)$ is the probability of performing action $a_t \in \mathcal{A}$ in state $s_t \in \mathcal{S}$, that maximizes the expected discounted reward collected by the agent $\mathcal{J}_r(\policy) = \mathbb{E}_{\sigma\sim \policy(\tau)}\sum^\infty_{k=0} \gamma^k r(s_{t+k},a_{t+k})$. We define a trajectory $\sigma = (s_0, a_0, \dots, s_T,a_T)$ of length $T$, and initial state $s_0$ following the state distribution $P_0(s_0)$.

The task of deriving a reward function $\rewardnetwork$ from human preferences, instantiated as a neural-network parameterized by $\psi$, can be defined as a supervised learning problem~\cite{christiano2017deep} where the goal is to predict reward signals from sequences of state-action pairs~\cite{wirth2017survey,christiano2017deep}. We define trajectory segments as a sub-sequence of the full trajectory $\sigma$~\cite{wilson2012bayesian}: the $j$-th segment, with length $m$, is $\segment^j=\left((\s^j_t,\action^j_t),\dots,(\s^j_{t+m-1},\action^j_{t+m-1})\right)$, capturing state-action pairs $(s_t, a_t)$ from time $t$ to $t+m$. Human users compare pairs of trajectory segments $(\segment^0,\segment^1)$ and assign a \emph{preference}, $\preference \in \{0,0.5,1\}$: $\preference=0$ implies favoring $\segment^0$ over $\segment^1$ (i.e., $\segment^0 \succ \segment^1$); $\preference=1$ suggests a preference for $\segment^1$ over $\segment^0$, (i.e., $\segment^1\! \succ \segment^0$); $\preference=0.5$ reflects an equal preference for both segments. The probability of a human preferring $\segment^0\! \succ \segment^1$ can be modeled to be exponentially dependent on the sum of rewards over the segments' length~\cite{bradley1952rank}:
\begin{equation}
\begin{split}
\!\!\!\!\!P_\psi[\segment^0\!\! \succ\!\! \segment^1] \!\!=\!\! \frac{\mathrm{exp}(\sum_{\substack{t}}\!\rewardnetwork(\s^0_t,\action^0_t))}{\mathrm{exp}(\sum_{\substack{t}}\!\rewardnetwork(\s^0_t,\action^0_t))\!+\!\mathrm{exp}(\sum_{\substack{t}}\!\rewardnetwork(\s^1_t,\action^1_t))}.
\label{eq:softmax}
\end{split}
\end{equation}
We train the reward function $\rewardnetwork$ as a binary classifier that predicts human preferences on new trajectory segments, serving as a proxy for the actual (unknown) reward function. The preferences gathered from human participants, along with the corresponding segments, are compiled into a labeled dataset $\dataset_l$, consisting of query triples $q= (\segment^0,\segment^1,\preference)$. Using this dataset we can optimise $\rewardnetwork$ by minimizing the binary cross-entropy loss:
\begin{equation}
\begin{split}
\loss_{\text{CE}}(\rewardnetwork,\dataset_l)  = - \mathbb{E}_{\substack{q \sim \dataset_l }} 
  & \Big[(1-\preference)\log P_\psi(\segment^0 \succ \segment^1) \\
  & + \preference \log P_\psi(\segment^1 \succ \segment^0)\Big].
\end{split}
\label{eq:cross}
\end{equation}

\section{Catastrophic Reward Forgetting\\ In Style Adaptation}
\label{sec:challenge}

Collecting human preferences for the style adaptation of robotic behavior is technically challenging and time-consuming. Due to this, standard datasets of human preferences for robotic adaptation often only consist of few hundreds to thousands of preference queries~\cite{christiano2017deep,lee2021pebble,hejna2023few}. In this small-data regime, adapting reward functions using standard PbRL algorithms often results in overfitting to the reward distribution of human preferences~\cite{zhang2018dissection,kannan2023deft}.

To highlight the problem of style adaptation in the low-data regime, in Figure~\ref{fig:visual} we consider the \texttt{Drawer-Close} simulation environment~\cite{yu2020meta}. We pre-train the policy and reward model $\rewardnetwork$ of the agent to be able to close the drawer (details regarding the experimental setup can be found in Section~\ref{sec:experiment}). As the style adaptation target we consider a human user that prefers that the robot closes the drawer from the right side: we collect 200 synthetic preferences that value trajectories that close the drawer from the right side over trajectories that do so from the left side.

In Figure~\ref{fig:visual} we show a t-SNE visualization of state-action pairs for this task, sampled from the original policy, and their respective reward value under the original and adapted reward model. We clearly observe that naively fine-tuning the reward model with small amounts of preference data results in a severe degradation of a large area of the state-action space, as high reward is given only to a narrow region of the state-action space (highlighted in dark green). A policy trained on the fine-tuning reward model effectively learns to avoid the left side of the drawer. However, it does so at the expense of the main goal of closing a drawer: for example, the robot might erroneously attempt to push the table from the right. We call this phenomena \textit{catastrophic reward forgetting} (CRF), where overfitting to a new style adaptation degrades the reward model's effectiveness and drastically undermines the performance on the overall task.

\section{Sample-efficient Style Adaptation via Low-Rank Matrices}
\label{sec:method}

\begin{figure*}[t]
	\centering
        \includegraphics[width=\linewidth, trim=20pt 11pt 20pt 5pt, clip]{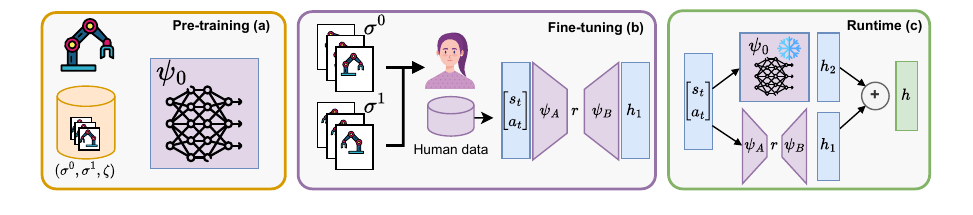}
        \caption{\textbf{The FLoRA framework.} (a) We pre-train a reward model with parameters $\psi_0$ by interleaving policy and reward training, following any preference-based RL algorithm; (b) Given new preferences from a human user, a new set of weights, $\psi_A$ and $\psi_B$, are introduced and fine-tuned while collecting novel feedback to adapt the network to a new task; (c) During run-time, the forward pass is made by summing both the frozen weights and the reward-adapted weights.}
 \vspace{-15pt}
	\label{fig:framework}
\end{figure*}

The results of Figure~\ref{fig:visual} highlight that human preferences occupy a small region of the full state-action space of the task in the low-data regime. Motivated by this result, and by recent work on the small \emph{intrinsic dimensionality} of reward models~\cite{aghajanyan2021intrinsic}, we explore the use of Low-Rank Adaptation (LoRA) for the style adaptation of pre-trained robotic behavior to human preferences. We contribute \emph{FLoRA}, a novel sample-efficient framework to adapt reward models with human preferences, agnostic to the underlying PbRL approach. As depicted in Figure~\ref{fig:framework}, FLoRA enables reward style adaptation without retraining the entire original reward model.

Consider a reward function $\rewardnetwork$, pre-trained on dataset $\dataset_l$ to minimize Equation~\ref{eq:cross} (Figure~\ref{fig:framework}a). Following~\cite{hu2021lora}, we extract the pre-trained reward function weight matrix $\psi_0\in \mathbb{R}^{d\times k}$, where $d$ and $k$ are the sizes of the input and the output respectively, for any given layer. Instead of directly fine-tuning $\psi_0$, which can be high-dimensional and prone to overfitting, we introduce a low-rank adaptation to modify it efficiently. Specifically, we decompose the weight update as follows: ${\psi_0+\Delta \psi=\psi_0+\psi_B\psi_A}$. Here, $\psi_B\in \mathbb{R}^{d\times r}$, $\psi_A\in \mathbb{R}^{r\times k}$, and the rank $r$ can be significantly smaller than $d$ or $k$.

% Consider a reward function $\rewardnetwork$, pre-trained on dataset $\dataset_l$ to minimize Equation~\ref{eq:cross} (Figure~\ref{fig:framework}a). Following~\cite{hu2021lora}, we extract the pre-trained reward function weight matrix $\psi_0\in \mathbb{R}^{d\times k}$, where $d$ and $k$ are the sizes of the input and the output respectively, for any given layer. The underlying goal of our method is not to update directly $\psi_0$, but instead employ a low-rank decomposition: ${\psi_0+\Delta \psi=\psi_0+\psi_B\psi_A}$. Here, $\psi_B\in \mathbb{R}^{d\times r}$, $\psi_A\in \mathbb{R}^{r\times k}$, and the rank $r$ can be significantly smaller than $d$ or $k$. 

% Instead of fine-tuning all parameters of the original reward model, this approach limits the update of the original reward distribution to a lower-dimensional set of parameters to obtain the style updated reward distribution. Given a set of human preferences, in the form of feedback triples $\zeta$ to $\dataset_{l_\text{new}}$, we proceed to fine-tune the newly introduced parameters $\psi_A, \psi_B$, while $\psi_0$ remains unchanged (Figure~\ref{fig:framework}b). We note that both $\psi_0$ and $\Delta \psi=\psi_B \psi_A$ interact with the same input, and their output vectors are combined in a coordinate-wise manner (Figure~\ref{fig:framework}c): the modified forward pass for a given concatenated input $c_t = \begin{bmatrix} s_t, a_t \end{bmatrix}^\top$ 
% is represented as follows:

Rather than updating all parameters of the original reward model, this approach restricts modifications to the lower-dimensional subspace defined by $\psi_B$ and $\psi_A$, effectively controlling the complexity of the adaptation. Given a set of human preferences in the form of feedback triples $\zeta$ to $\dataset_{l_\text{new}}$, we fine-tune only the newly introduced parameters $\psi_A$ and $\psi_B$, while keeping $\psi_0$ unchanged (Figure~\ref{fig:framework}b). Notably, both $\psi_0$ and $\Delta \psi = \psi_B \psi_A$ interact with the same input, and their output vectors are combined coordinate-wise (Figure~\ref{fig:framework}c). The modified forward pass for a given concatenated input $c_t = \begin{bmatrix} s_t, a_t \end{bmatrix}^\top$ is given by:
\begin{equation}
h = \psi_0 \begin{bmatrix} s_t, \\ a_t \end{bmatrix} + \Delta \psi \begin{bmatrix} s_t, \\ a_t \end{bmatrix} = \psi_0 \begin{bmatrix} s_t,\\ a_t \end{bmatrix} + \psi_B\psi_A \begin{bmatrix} s_t,\\ a_t \end{bmatrix}.
\label{eq:lora}
\end{equation}
The initialization process involves setting $\psi_A$ to a random Gaussian distribution, which injects diversity into the adaptation process, while $\psi_B$ is initialized to zero, ensuring that ${\Delta \psi=\psi_B\psi_A}$ starts as zero at the onset of training. Following \cite{hu2021lora}, the term $\Delta \psi \begin{bmatrix} s_t & a_t \end{bmatrix}^\top$ is scaled by a factor of $\alpha/r$, where $\alpha$ is a constant related to $r$. In the context of optimization using ADAM~\cite{kingma2014adam}, adjusting $\alpha$ effectively modulates the step size of updates, making it analogous to tuning the learning rate.

Applying our low-rank decomposition to the problem of style adaptation in the low-preference-data regime offers significant advantages: i) it introduces no additional inference latency at runtime since computations can be performed in parallel, which is particularly important for control tasks; ii) it enables efficient training by reducing the number of trainable parameters, thereby lowering the risk of overfitting; (iii) our framework is naively compatible with other techniques and is agnostic to the underlying PbRL algorithm used.

\section{Evaluation}
\label{sec:experiment}

Our evaluation highlights the performance of FLoRA in style adaptation of robotic behavior accordingly to the following two questions:
\begin{enumerate}
    \item[\textbf{Q1}:] Does FLoRA enable sample-efficient style adaptation of robotic behavior to human preferences while mitigating CRF?
    \item [\textbf{Q2}:] Can we employ FLoRA to efficiently adapt real-world robot behavior with human feedback?
\end{enumerate}
For more details on the experimental setup, training hyperparameters and human preferences please refer to Appendix~\ref{appendix:implementation_details}.

\nopar{\textbf{Scenarios} \quad We benchmark FLoRA on literature-standard simulation tasks and real-world tasks: from the DeepMind Control Suite~\cite{tunyasuvunakool2020dm_control}, we consider the \texttt{Walker}, \texttt{Cheetah}, and \texttt{Quadruped Walker} tasks; from the Meta-World robotic benchmark~\cite{yu2020meta}, we consider the \texttt{Button-Press}, \texttt{Button-Press-Wall}, and \texttt{Drawer-Close} tasks with high-dimensional input observations. For the real-world experiments, we consider two different robotic tasks: a manipulation task (\texttt{RW-Drawer-Close}) and a navigation task (\texttt{Spot-Follow}). In the \texttt{RW-Drawer-Close} task, we use a 7-DoF robotic manipulator to close a drawer. In the \texttt{Spot-Follow} task, we train a four-legged mobile robot to follow the human user around an environment without colliding with the human.}

\nopar{\textbf{Baselines} \quad We compare FLoRA against three baseline frameworks to adapt robotic behavior: \emph{fine-tuning} the complete reward model to novel human preferences, an approach akin to~\cite{hejna2023few}; \emph{SURF}~\cite{park2022surf}, a recently proposed semi-supervised data augmentation-based method suitable for adaptation; and \emph{Co-training}, a baseline where we assume we have access to the original dataset used to pretrained the reward model to fine-tune the model along the human preferences, similarly to DAGGER~\cite{ross2011reduction}. To learn policies, we use \emph{PEBBLE}~\cite{lee2021pebble}, a state-of-the-art preference-based RL algorithm: the reward function is an MLP with three hidden layers of size $\{256,256,256\}$; and the underlying RL algorithm is based on SAC~\cite{haarnoja2018soft} with the default parameters found in~\cite{lee2021pebble}. For FLoRA we performed minimal hyperparameter tuning and considered $r=16$ and a relative weighting of $\alpha/r=1$.}

\begin{figure*}[t]
	\centering
	\includegraphics[width=0.99\linewidth]{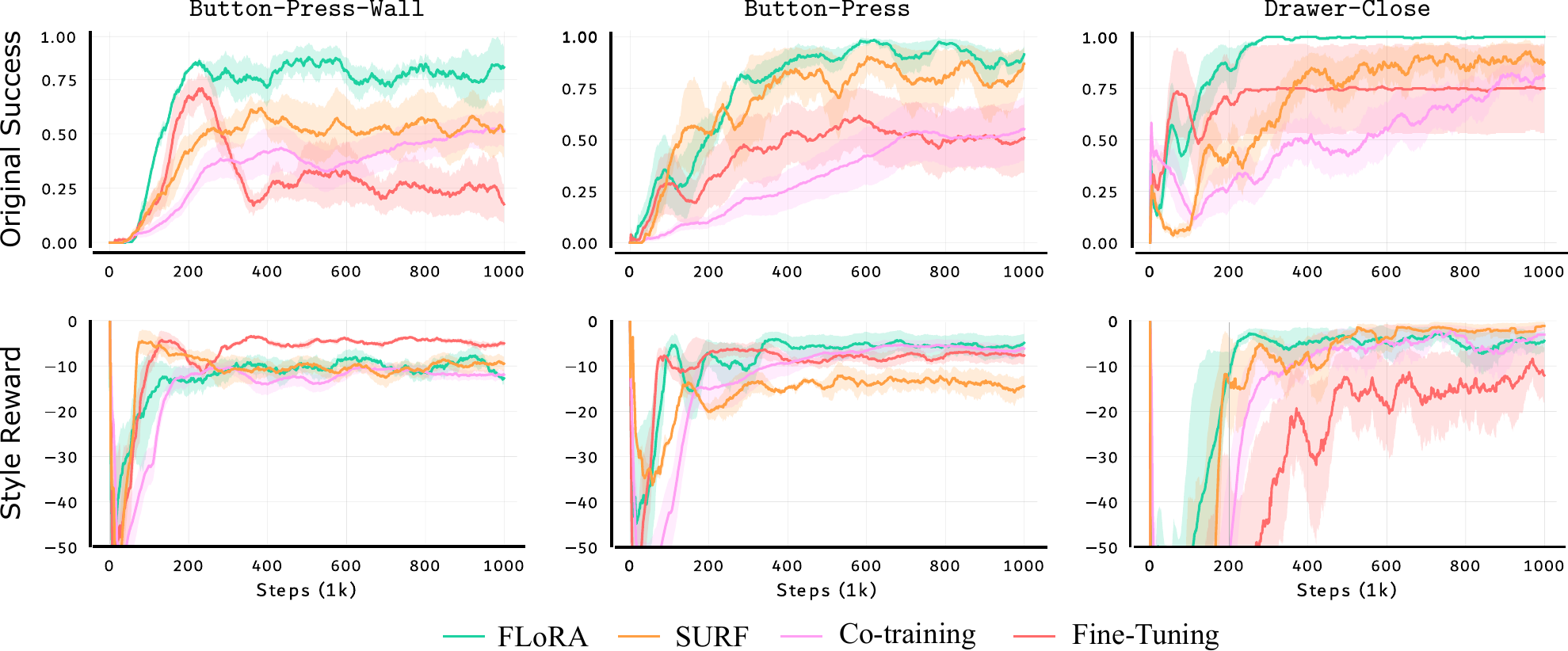}
	\caption{\textbf{Training Curves on Meta-World Environments.} (Top) Success rate of the different methods in the original task; (Bottom) Returns of the different methods measured with the style adaptation reward function. We report the mean (solid lines) and the standard error (shaded area) of the performance, averaged over 5 randomly selected seeds. Higher is better.}
 \vspace{-15pt}
	\label{fig:meta-rewards}
\end{figure*}

\nopar{\textbf{Style Preferences} \quad In simulation environments, we collect synthetic human preferences based on the cumulative sum of rewards from different agent trajectories with respect to the original/style reward model, using an error rate of approximately $10\%$ and trajectory segments of size $50$, as done in similar works~\cite{christiano2017deep,lee2021pebble,hejna2023few}. To collect novel human preferences in each scenario, we design custom style reward functions for each environment, representing a new additional objective used to label the new queries. For the real-world tasks, we performed style adaptation using two different robots: a quadruped mobile robot (Boston Dynamics' Spot) and a 7-DoF manipulator (Franka). The mobile robot's main objective is to follow human users without colliding with them, and the adapted style is for the robot to stay in the positive x-axis direction relative to the human while following them. The manipulator robot's main objective is to successfully close a drawer, and in the adapted style, it closes the drawer slowly. In both tasks, the queries are answered through a synthetic oracle, and in this case we observed no difference compared to being answered by the authors.}

\section{Results}
\label{sec:results}

We present the main results of the benchmark evaluation in Table~\ref{tab:main_results} and we show the training curves of the different methods in Figure~\ref{fig:meta-rewards}. \footnote{We include additional results in Appendix~\ref{appendix:additional_results}, including training curves, ablation studies on $r$, $\alpha$ and how we can naively combine FLoRA with other adaptation methods.}

\subsection{FLoRA enables style adaptation while mitigating CRF}
\label{sec:results:q1}
The results in Table~\ref{tab:main_results} highlight how FLoRA mitigates CRF, especially in the challenging Meta-World environment, \texttt{Drawer-Close}, by achieving a higher combined (original + style) reward compared to the baselines, (on average 19.46\% higher). This demonstrates that our method effectively adapts policy styles to novel human preferences while maintaining the ability to perform the original task. 

\begin{table*}[t]
    \centering
    \caption{
        \textbf{Style Adaptation in Benchmark Environments.} We report the returns achieved by different adaptation methods in regards to the original and the style-adapted reward functions. All results are averaged over 5 randomly selected seeds. We highlight the best average results in \colorbox{HighlightBest}{blue} and the worst average result in \colorbox{HighlightWorst}{orange}. Higher is better.}
    \label{tab:main_results}
    \vspace{4pt}
    \renewcommand{\arraystretch}{1.25} % Adjust row spacing
    \setlength\arrayrulewidth{0.05pt} % Adjust width of dashed lines
    \resizebox{0.7\textwidth}{!}{
    \begin{tabular}{l l l c c c c c}
        \toprule
        \textbf{Environment} & $|\mathcal{D}_l|$ & \textbf{Reward} & \textbf{FLoRA} & \textbf{SURF} & \textbf{Co-training} & \textbf{Fine-tuning} \\
        \midrule
        \multirow{4}{*}{\texttt{Drawer-Close}} & 
        
         \multirow{2}{*}{200} & 
        Original & \entry{\colorbox{HighlightBest}{3576}}{927} &  \entry{3506}{581} & \entry{\colorbox{HighlightWorst}{2432}}{1012} & \entry{2971}{854} \\
         & &  Style & \entry{-94}{41} & \entry{\colorbox{HighlightWorst}{-126}}{21} & \entry{-29}{20} & \entry{\colorbox{HighlightBest}{-18}}{15} \\
         &  \multirow{2}{*}{400} & Original & \entry{\colorbox{HighlightBest}{4622}}{59} & \entry{4203}{403} & \entry{3663}{361} &  \entry{\colorbox{HighlightWorst}{3412}}{991} \\ 
        & & Style & \entry{-3.87}{2.1} & \entry{\colorbox{HighlightBest}{-1.04}}{0.84} & \entry{-3.01}{2.27} & \entry{\colorbox{HighlightWorst}{-10.65}}{5.38} \\ 
        \arrayrulecolor{gray!20}\hdashline\arrayrulecolor{black}
        \multirow{4}{*}{\texttt{Button-Press}} & 
         
         \multirow{2}{*}{200} & Original & \entry{\colorbox{HighlightBest}{1044}}{449} & \entry{738}{477} & \entry{1035}{339} & \entry{\colorbox{HighlightWorst}{210}}{45} \\ 
        & & Style & \entry{-13}{2.69} & \entry{-21}{9.33} & \entry{\colorbox{HighlightBest}{-4.88}}{1.84} & \entry{\colorbox{HighlightWorst}{-22}}{6.13} \\ 
        & \multirow{2}{*}{400} & 
        Original & \entry{1825}{661} & \entry{\colorbox{HighlightBest}{1941}}{631} & \entry{1638}{304} & \entry{\colorbox{HighlightWorst}{1290}}{512} \\
         & &  Style & \entry{\colorbox{HighlightBest}{-4.40}}{1.81} & \entry{\colorbox{HighlightWorst}{-13.40}}{2.19} & \entry{-11.93}{1.11} & \entry{-7.41}{1.77} \\
        \arrayrulecolor{gray!20}\hdashline\arrayrulecolor{black}
        \multirow{4}{*}{\texttt{Button-Press-Wall}} & 
        
         \multirow{2}{*}{200} & 
        Original & \entry{1134}{449} & \entry{\colorbox{HighlightBest}{1258}}{80} & \entry{952}{234} & \entry{\colorbox{HighlightWorst}{393}}{78} \\
         & &  Style & \entry{-14}{1.27} & \entry{-16}{1.96} & \entry{\colorbox{HighlightBest}{-6.64}}{1.88} & \entry{\colorbox{HighlightWorst}{-18}}{3.87} \\
         &  \multirow{2}{*}{400} & Original & \entry{\colorbox{HighlightBest}{1645}}{628} & \entry{1585}{365} & \entry{1392}{403} & \entry{\colorbox{HighlightWorst}{727}}{219} \\ 
        & & Style & \entry{\colorbox{HighlightWorst}{-13.40}}{12.3} & \entry{-8.69}{2.00} & \entry{-6.11}{1.50} & \entry{\colorbox{HighlightBest}{-4.87}}{0.73} \\
        \arrayrulecolor{gray!20}\hdashline\arrayrulecolor{black}
        \multirow{4}{*}{\texttt{Walker}} & 
        
         \multirow{2}{*}{50} & 
        Original & \entry{482}{243} & \entry{\colorbox{HighlightBest}{577}}{105} & \entry{561}{91} & \entry{\colorbox{HighlightWorst}{328}}{144} \\
         & &  Style & \entry{\colorbox{HighlightWorst}{633}}{213} & \entry{775}{105} & \entry{\colorbox{HighlightBest}{826}}{23} & \entry{722}{112} \\
         &  \multirow{2}{*}{100} & Original & \entry{\colorbox{HighlightBest}{686}}{75} & \entry{632}{\phantom{1}38} & \entry{622}{96} & \entry{\colorbox{HighlightWorst}{333}}{138} \\ 
        & & Style & \entry{\colorbox{HighlightBest}{917}}{\phantom{1}46} & \entry{\colorbox{HighlightWorst}{856}}{\phantom{1}10} & \entry{916}{\phantom{1}8} & \entry{864}{\phantom{1}71} \\ 
        \arrayrulecolor{gray!20}\hdashline\arrayrulecolor{black}
        \multirow{4}{*}{\texttt{Cheetah}} & 
        
         \multirow{2}{*}{20} &  Original &
        \entry{\colorbox{HighlightBest}{429}}{98} & \entry{226}{65} & \entry{294}{111} & \entry{\colorbox{HighlightWorst}{210}}{50} \\ 
        & & Style & \entry{-92}{43} & \entry{\colorbox{HighlightBest}{-49}}{16} & \entry{\colorbox{HighlightWorst}{-116}}{77} & \entry{-56}{65} \\
        & \multirow{2}{*}{50} &  Original &
        \entry{\colorbox{HighlightBest}{450}}{52} & \entry{450}{60}* & \entry{386}{40} & \entry{\colorbox{HighlightWorst}{319}}{107} \\ 
        & & Style & 
        \entry{-41}{23} & \entry{-38}{10} & \entry{\colorbox{HighlightBest}{-25}}{16} & \entry{\colorbox{HighlightWorst}{-45}}{16}\\ 
        \arrayrulecolor{gray!20}\hdashline\arrayrulecolor{black}
        \multirow{4}{*}{\texttt{Quadruped}} & 
        
        \multirow{2}{*}{50} & Original &
        \entry{\colorbox{HighlightBest}{166}}{37} & \entry{146}{55} & \entry{123}{26} & \entry{\colorbox{HighlightWorst}{99}}{57} \\ 
        & & Style & \entry{861}{7} & \entry{\colorbox{HighlightBest}{863}}{6} & \entry{860}{8} & \entry{\colorbox{HighlightWorst}{806}}{55}\\
        & \multirow{2}{*}{100} & Original & 
        \entry{\colorbox{HighlightBest}{191}}{55} & \entry{187}{46} & \entry{131}{33} & \entry{\colorbox{HighlightWorst}{118}}{41} \\ 
        & & Style &
        \entry{870}{6} & \entry{\colorbox{HighlightBest}{873}}{7} & \entry{867}{2} & \entry{\colorbox{HighlightWorst}{844}}{43} \\  
        \bottomrule
    \end{tabular}}
    \vspace{-10pt}
\end{table*}
FLoRA outperforms the standard fine-tuning baseline in adapting to human preferences, demonstrating greater sample efficiency while maintaining the ability to perform the original task. This baseline consistently suffers from CRF, attaining the worst performance in the original task across the majority of the environments. FLoRA either outperforms or performs on par with the data augmentation baseline (SURF). Our method outperforms SURF in two of the three higher-dimensional complex robot tasks, such as the Meta-World robotics tasks. We show that FLoRA leads to a larger overall increase in both original and style rewards compared to SURF~\cite{park2022surf}, achieving an average increase of 21.24\% with the same number of feedback samples. This result shows that our approach is able to scale to more complex scenarios using only a small number of human preferences, without requiring additional synthetic preferences. Our method also outperforms or performs on par with Co-training. This baseline model assumes access to the original training dataset, which may be an unrealistic assumption when adapting pretrained robotic behavior. FLoRA, instead, only employs the novel human preferences for the adaptation.

In Figure~\ref{fig:meta-rewards} we highlight that the consequences of a steep performance drop in the original reward function often mean failing to perform basic tasks (e.g., reaching the button in the \texttt{Button-Press} environment), resulting in the inability of the agent to adapt to the human preference: the fine-tuning approach often achieves the worst performing result in both original and style returns. Instead, FLoRA maintains high levels of task completion rates across all environments.

\begin{figure}[t]
    \centering
    \begin{minipage}[t]{0.1\columnwidth}
        \centering
        \vspace{16pt} % Aligns the top of the minipage with the top of the row
        \includegraphics[height=0.049\textheight]{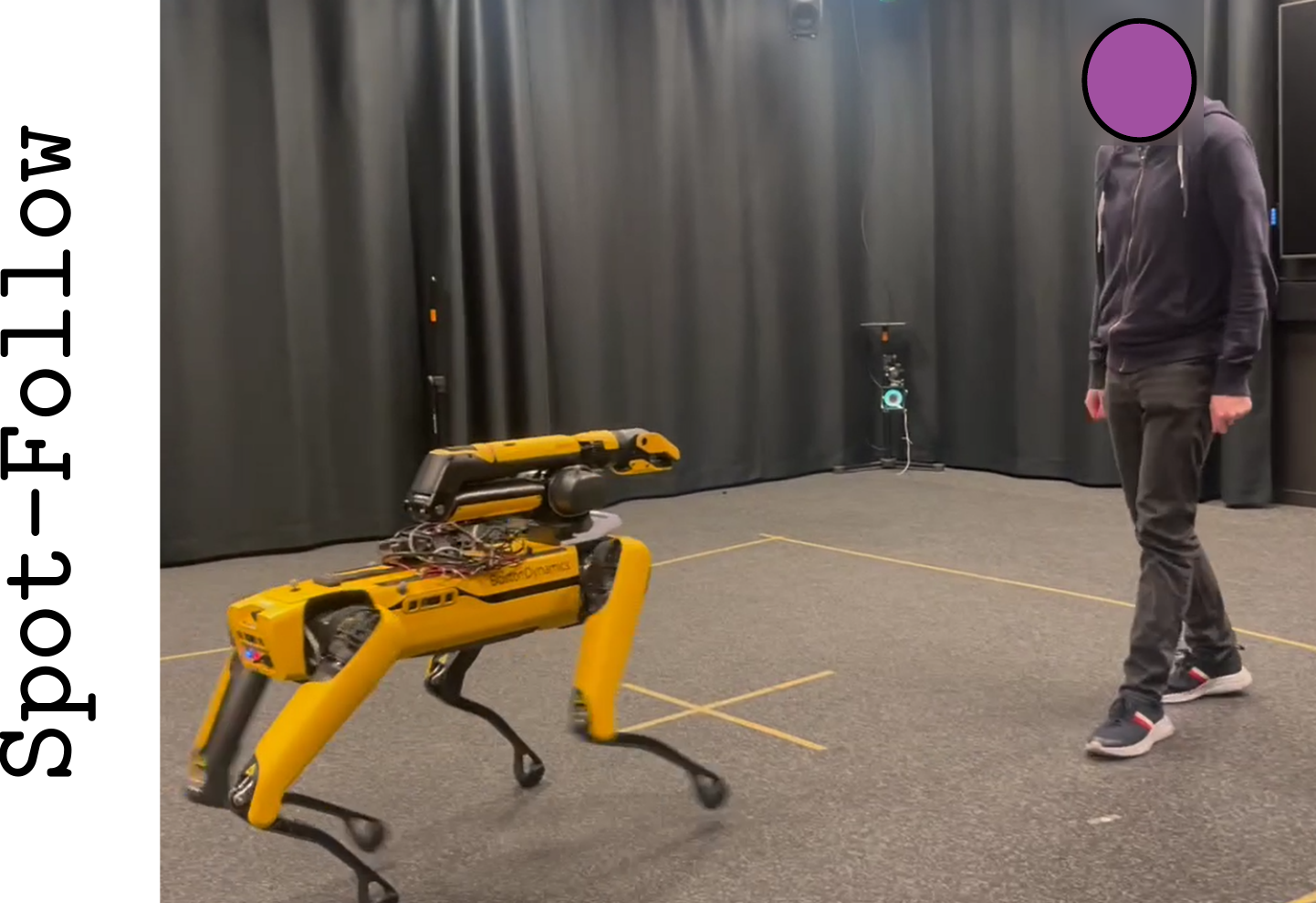}\\
        \vspace{12pt} % Adjust the space between the images as needed
        \includegraphics[height=0.049\textheight]{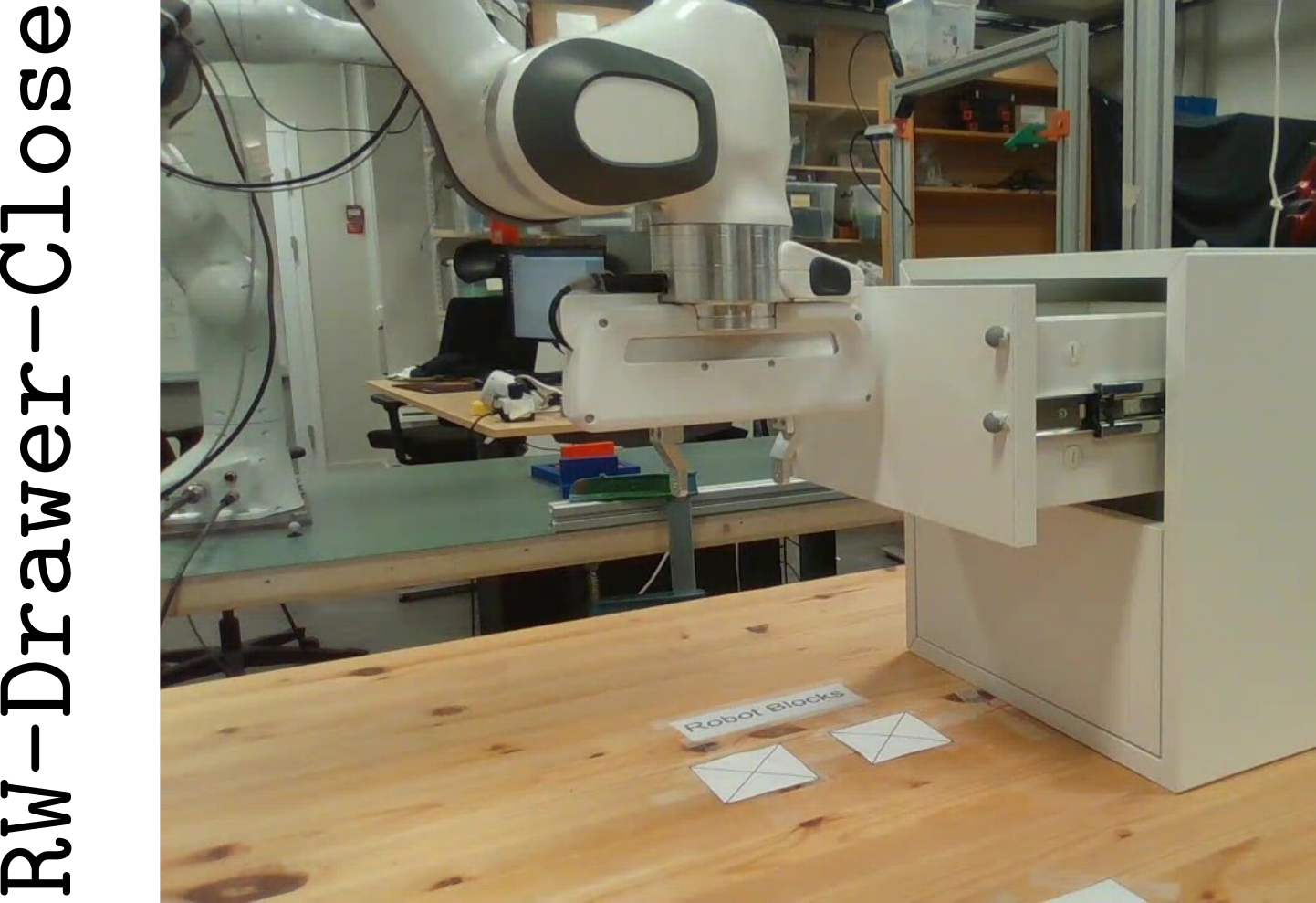}
    \end{minipage}%
    \hspace{23pt}
    \begin{minipage}[t]{0.78\columnwidth}
        \centering
        \vspace{0pt} % Aligns the top of the minipage with the top of the row
        \renewcommand{\arraystretch}{1.4} % Adjust row spacing
        \setlength\arrayrulewidth{0.05pt} % Adjust width of dashed lines
        \resizebox{1.0\textwidth}{!}{
        \begin{tabular}{l c c c}
            \toprule
            \textbf{Metrics} & \textbf{FLoRA} & \textbf{SURF} & \textbf{Fine-tuning} \\
            \midrule
            Success Rate $(\%, \uparrow)$  & 65 & 75 & 10
            \\
            Near Collision Rate $(\%, \downarrow)$  & 10 & 40 & 20 
            \\
            Average Finish Distance $(m, \downarrow)$  & 0.59 & 1.13 & 0.61 
            \\
            \arrayrulecolor{gray!20}\hdashline\arrayrulecolor{black}
            \vspace{-6pt}
            \\
            \textbf{Metrics} & \textbf{FLoRA} & \textbf{SURF} & \textbf{Fine-tuning} \\
            \midrule\arrayrulecolor{gray!20}\hdashline\arrayrulecolor{black}
            Distance Pushed $(cm, \uparrow)$  & 2.58 & 1.40 & 0.05
            \\
            Velocity $(ms^{-1}, \downarrow)$  & 0.018 & 0.029 & 0.023
            \\
            \bottomrule
        \end{tabular}
        }
    \end{minipage}
    % \vspace{-10pt}
    \caption{\textbf{Style Adaptation in Real-World Tasks.} For the \texttt{Spot-Follow} task, we report the task success rate (being able to complete the original task), the near-collision rate (counted when Spot overrides the policy with collision avoidance), and the average final distance from the goal, averaged over 20 episodes. For the \texttt{RW-Drawer-Close} task, we report how far it pushed in the drawer and the velocity of the end-effector,  averaged over 6 episodes. The arrows indicate the direction of improvement.}
    \label{fig:combined}
    \vspace{-10pt}
\end{figure}

\subsection{FLoRA enables efficient style adaptation of real-world robotic behavior}
\label{sec:results:q3}

\nopar{\textbf{Manipulation Task} \quad We present the results of the \texttt{RW-Close-Drawer} task in Figure~\ref{fig:combined}. The results show that FLoRA can both close the drawer further (main task) than both SURF and Fine-Tuning and at a lower speed (style) across all methods. None of the methods were able to completely close the drawer, due to the significant \emph{sim-to-real gap} between the physics of the spring mechanism in the real-world drawer and the simulated one.
}

\nopar{\textbf{Navigation Task} \quad We present the results of the \texttt{Spot-Follow} task in Figure~\ref{fig:combined}. The results show that FLoRA can adapt to a new style without losing the ability to perform the original task: being able to follow the human (shown by the lower average final distance to the goal) while having the lowest near collision rate of 10\%. Both SURF and fine-tuning methods exhibit signs of catastrophic reward forgetting: SURF exhibits much higher near-collision rates with humans, and simple fine-tuning shows drastically lower success rates and incorrect final positions of the robot.}

\section{Discussion and Limitations}
\label{sec:conclusion}

\nopar{\textbf{Connection to control foundation models} \quad We create FLoRA with the future landscape of robotics in mind. Similar to the emergence of foundation models~\cite{bommasani2021opportunities}, such as large language models (LLMs)~\cite{devlin2018bert,brown2020language} and vision-language models (VLMs)~\cite{jia2021scaling}, we are beginning to observe the rise of foundation robot control policies~\cite{padalkar2023open,gdm2024autort,team2024octo}. Our work is a first step in demonstrating the feasibility of stylistically adapting reward functions to the personalized preferences of humans. In future work, we would like to explore directly aligning policies with preferences via low-rank adaptation.}

\nopar{\textbf{FLoRA for multiple style adaptation} \quad For future work we also aim to explore using multiple LoRAs for style adaptation in robotic control tasks. Recent advancements in combining multiple LoRAs for text~\cite{shah2023ziplora} and image~\cite{zhong2024multi} generation indicate promising potential for this approach. However, robotic control tasks present unique challenges, as failure to adhere to the original task can lead to unsafe behaviors that abruptly terminate policy execution.}

 \nopar{\textbf{Limitations} \quad The limitations of FLoRA share commonalities with those of LoRAs in general. While scalability is not an issue, as LoRAs are designed for arbitrarily large models, they are intended to adapt them. If the models require significant parameter changes or must be completely re-trained, LoRAs may not be sufficient, even for high-rank $r$ values. Additionally, the optimal values of $r$ may be environment-dependent.}

\section*{ACKNOWLEDGMENT}

This work was partially funded by grants from the Swedish Research Council (2024-05867), the Swedish Foundation for Strategic Research (SSF FFL18-0199), the Digital Futures research center, the Vinnova Competence Center for Trustworthy Edge Computing Systems and Applications at KTH, and the Wallenberg Al, Autonomous Systems and Software Program (WASP) funded by the Knut and Alice Wallenberg Foundation.  Additionally, this work was partially supported by the HORIZON-CL4-2021-HUMAN-01 ELSA project.

% This research has been carried out as part of the Vinnova Competence Center for Trustworthy Edge Computing Systems and Applications at KTH, and partially supported by the Swedish Foundation for Strategic Research (SSF FFL18-0199) and the Wallenberg AI, Autonomous Systems  and  Software  Program (WASP) funded by the Knut and Alice Wallenberg Foundation. Additionally, this work was partially supported by the HORIZON-CL4-2021-HUMAN-01 ELSA project.

%%%%%%%%%%%%%%%%%%%%%%%%%%%%%%%%%%%%%%%%%%%%%%%%%%%%%%%%%%%%%%%%%%%%%%%%%%%%%%%%

\bibliography{main}
\bibliographystyle{IEEEtran}

\clearpage
\appendix

\subsection{Additional Results}
\label{appendix:additional_results}

\begin{figure*}[t]
    \centering
    \includegraphics[width=0.99\linewidth]{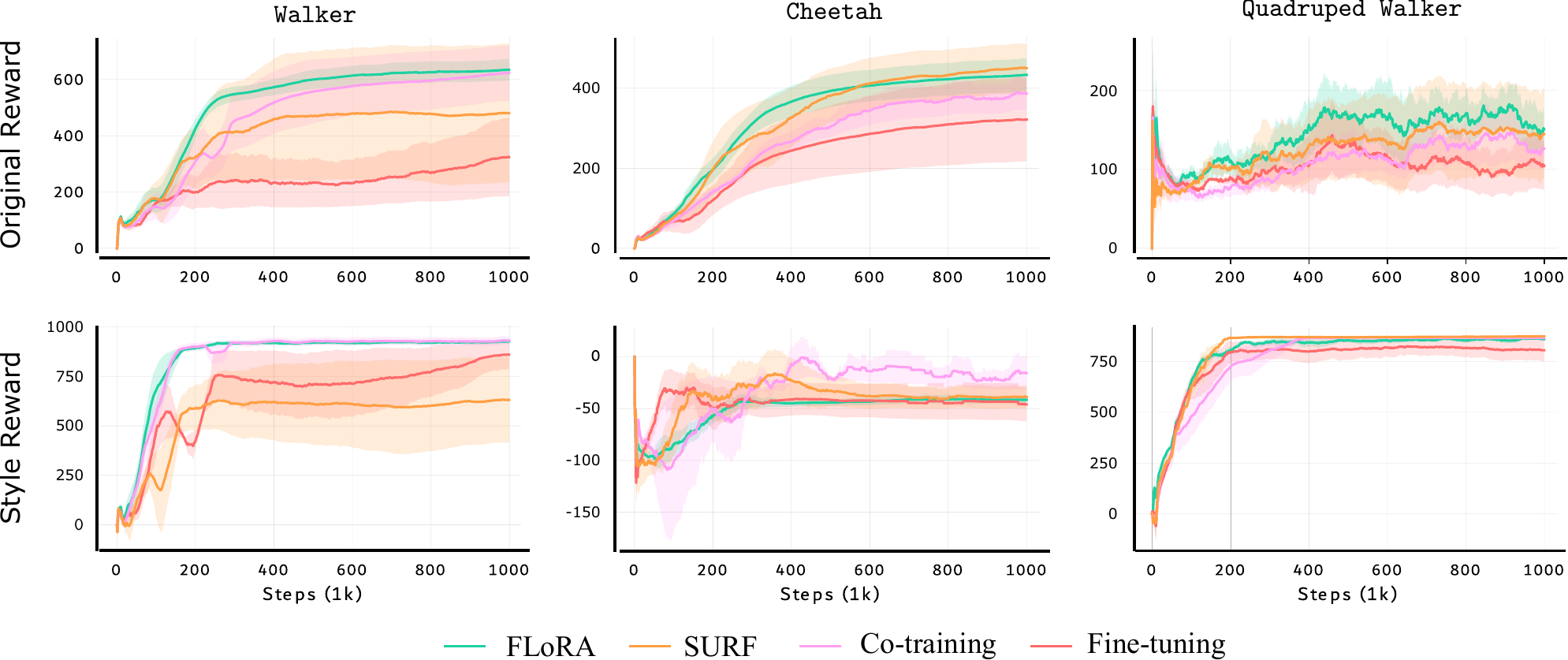}
    \caption{\textbf{Training Curves on DM-control environments.} (Top) Success rate of the different methods in the original task; (Bottom) Returns of the different methods measured with the style adaptation reward function. We report the mean (solid lines) and the standard error (shaded area) of the performance, averaged over 5 randomly selected seeds. Higher is better.}
    \label{fig:dm-rewards}
\end{figure*}

\subsubsection{DeepMind Control Suite Training Curves}
\label{appendix:additional_results_dm}

In Figure~\ref{fig:dm-rewards} we show the training curves for the DM Control environments \texttt{Walker}, \texttt{Cheetah} and \texttt{Quadruped-Walker}. We start from a pre-trained reward function across all conditions. Overall, FLoRA outperforms or performs on par with all baselines in maintaining the original reward while still adapting successfully to the style adaptation. Similarly to the Meta-World experiments, Fine-Tuning exhibits clear signs of CRF. For example in \texttt{Walker} overfitting to the style behaviour, considerably degrades the overall performance of the original task, which makes it also counter-productive to adapt to the stylized behaviour of jumping higher.

\subsubsection{FLoRA enhances adaptation frameworks by mitigating CRF}
\label{appendix:additional_results_crf}

\begin{table}[t]
    \centering
    \caption{
         \textbf{Style Adaptation in Benchmark Environments.} We report the returns achieved by SURF and our combination method ``FLoRA + SURF'' in regards to the original and the style-adapted reward functions. All results are averaged over 5 randomly selected seeds. We highlight in bold the best average results. Higher is better.}
    \label{tab:main_results}
    \vspace{4pt}
    \renewcommand{\arraystretch}{1.25} % Adjust row spacing
    \setlength\arrayrulewidth{0.05pt} % Adjust width of dashed lines
    \resizebox{0.48\textwidth}{!}{
    \begin{tabular}{l l l c c c}
        \toprule
        \textbf{Environment} & \textbf{Preferences} & \textbf{Reward} & \textbf{ FLoRA + SURF} & \textbf{SURF} \\
        \midrule
        \multirow{4}{*}{\texttt{Drawer-Close}} & 
         \multirow{2}{*}{200} & 
        Original & \entry[bold]{3945}{592} & \entry{3506}{581} \\
         & &  Style & \entry[bold]{-81}{24} & \entry{-126}{21} \\
         &  \multirow{2}{*}{400} & Original & \entry[bold]{4691}{90} & \entry{4203}{403} \\ 
        & & Style & \entry[bold]{-0.48}{0.84} & \entry{-1.04}{0.844} \\ 
        \arrayrulecolor{gray!20}\hdashline\arrayrulecolor{black}
        \multirow{4}{*}{\texttt{Button-Press}} & 
         
         \multirow{2}{*}{200} & Original & \entry[bold]{1284}{365} & \entry{738}{477} \\ 
        & & Style & \entry[bold]{-7.9}{1.77} & \entry{-21}{9.33} \\ 
        & \multirow{2}{*}{400} & 
        Original & \entry[bold]{2489}{525} & \entry{1941}{631} \\
         & &  Style & \entry[bold]{-5.2}{1.89} & \entry{-13.40}{2.19} \\
        \arrayrulecolor{gray!20}\hdashline\arrayrulecolor{black}
        \multirow{4}{*}{\texttt{Button-Press-Wall}} & 
         \multirow{2}{*}{200} & 
        Original & \entry[bold]{1557}{260} & \entry{1258}{80} \\
         & &  Style & \entry[bold]{-13}{1.97} & \entry{-16}{1.96} \\
         &  \multirow{2}{*}{400} & Original & \entry[bold]{1849}{538} & \entry{1585}{365} \\ 
        & & Style & \entry{-9.85}{2.09} & \entry[bold]{-8.69}{2.00} \\
        \arrayrulecolor{gray!20}\hdashline\arrayrulecolor{black}
        \multirow{4}{*}{\texttt{Walker}} & 
         \multirow{2}{*}{50} & 
        Original & \entry[bold]{721}{80} & \entry{577}{105} \\
         & &  Style & \entry[bold]{837}{51} & \entry{775}{105} \\
         &  \multirow{2}{*}{100} & Original & \entry[bold]{759}{79} & \entry{632}{38} \\ 
        & & Style & \entry[bold]{912}{8} & \entry{856}{10} \\
        \arrayrulecolor{gray!20}\hdashline\arrayrulecolor{black}
        \multirow{4}{*}{\texttt{Cheetah}} & 
         \multirow{2}{*}{20} &  Original &
        \entry[bold]{487}{33} & \entry{226}{65} \\ 
        & & Style & \entry[bold]{-32}{6} & \entry{-49}{16} \\
        & \multirow{2}{*}{50} &  Original &
        \entry[bold]{546}{29} & \entry{450}{60} \\ 
        & & Style & 
        \entry[bold]{-19}{17} & \entry{-38}{10} \\
        \arrayrulecolor{gray!20}\hdashline\arrayrulecolor{black}
        \multirow{4}{*}{\texttt{Quadruped}} & 
        \multirow{2}{*}{50} & Original &
        \entry[bold]{167}{24} & \entry{146}{55} \\ 
        & & Style & \entry[bold]{874}{4} & \entry{863}{6} \\
        & \multirow{2}{*}{100} & Original & 
        \entry[bold]{209}{57} & \entry{187}{46} \\ 
        & & Style &
        \entry[bold]{875}{5} & \entry{873}{7} \\  
        \bottomrule
    \end{tabular}}
\end{table}

We introduce an additional method that combines the data augmentation process of SURF with the low-rank adaptation process of our method, which we denote as ``FLoRA + SURF'' in Table~\ref{tab:main_results}. The results show that this combined method significantly improves the performance of SURF in the original task, mitigating CRF. Moreover, it highlights the versatility of our method, as it can be naively employed alongisde other approaches.

\subsubsection{Rank sensitivity}
\label{sec:sensitivity}
\begin{figure}[t]
    \centering
\includegraphics[width=0.99\linewidth]{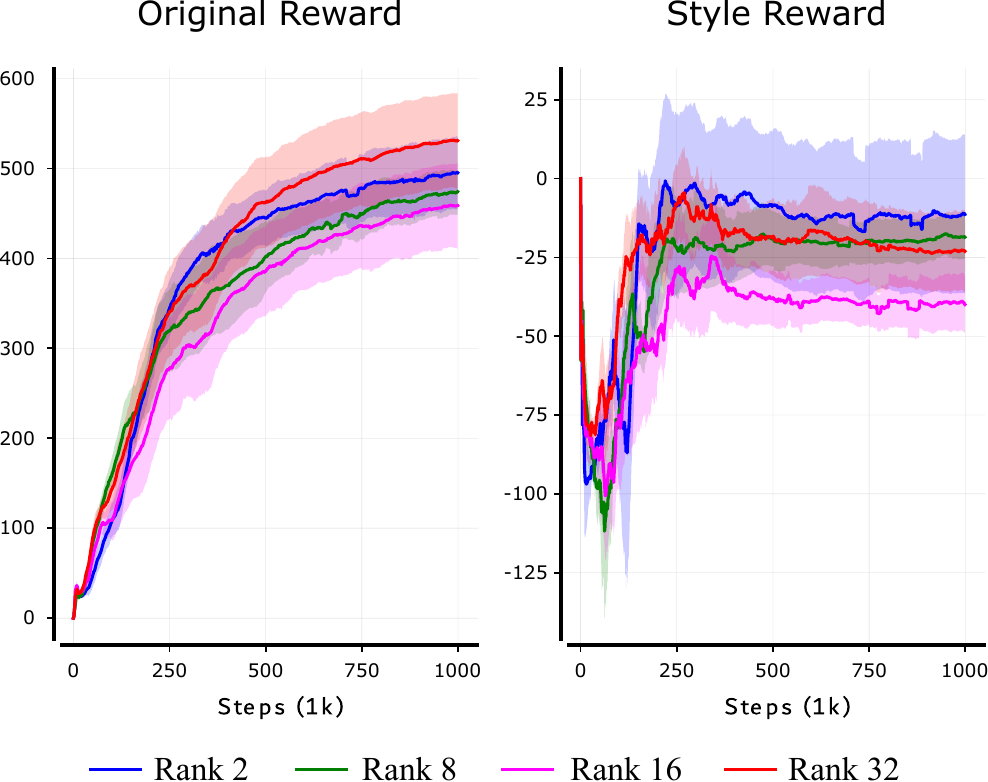}
    \vspace{-10pt}
    \caption{Ablation study of the rank of the LoRAs employed in our method, with $r=\{2,8,16, 32\}$, in the \texttt{Cheetah} scenario. Results averaged over five randomly-seeded runs.}
    \label{fig:ranks}

\end{figure}

We evaluate the sensitivity of FLoRA to the choice of rank size $r$ when adapting the reward function. We consider the FLoRA baseline, and conduct an ablation study, the results of which are presented in Figure~\ref{fig:ranks}. Generally, the style adaptation appears to be quite resilient to changes in rank. For example, if we compare against the results observed in the remaining conditions for the \texttt{Cheetah} (see Figure~\ref{fig:dm-rewards}), FLoRA performs compared with every other condition regardless of the rank. However, the selection of an optimal rank for FLoRA is inherently conditioned on both the complexity of the reward function and the environment. We leave the automatic setting of the value of $r$ for future work.

\begin{figure}[h]
    \centering
    \includegraphics[width=0.99\linewidth]{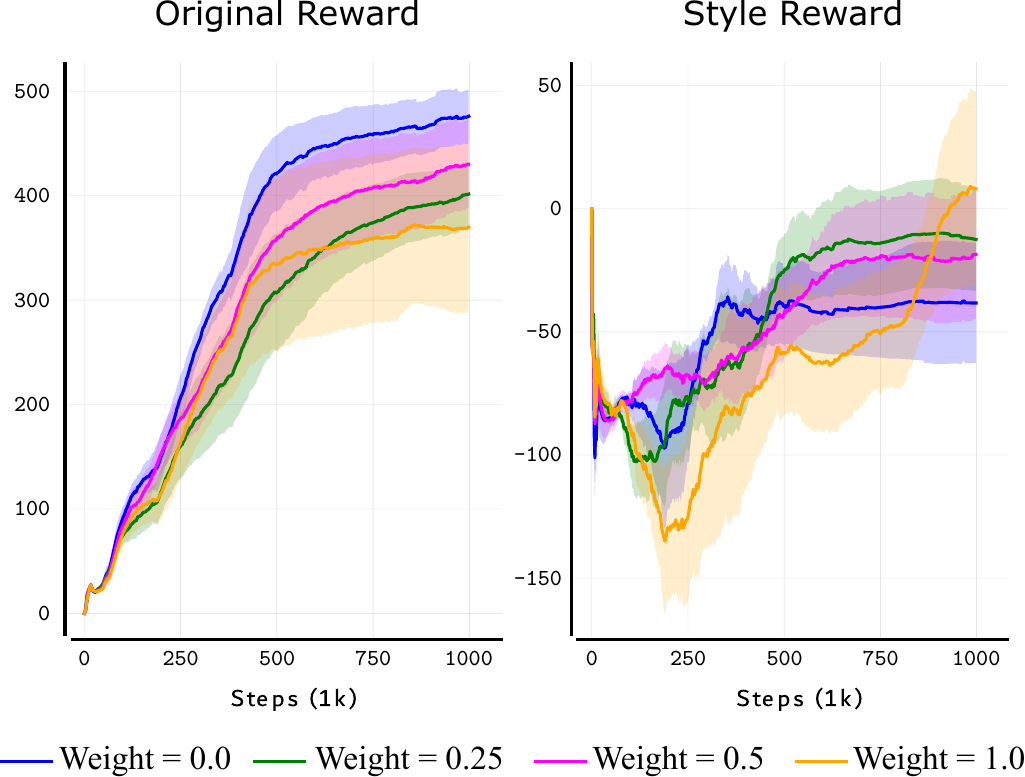}
    \vspace{-10pt}
    \caption{Ablation study of the weight $\alpha=\{0,0.25,0.5, 1.0\}$ during training of the LoRAs for our method in the \texttt{Cheetah} scenario. Results averaged over five randomly-seeded runs. }
    \label{fig:weights}
\end{figure}

\subsubsection{LoRA Weighting ($\alpha/r$) Sensitivity}
\label{sec:alphaablate}
We conduct an additional ablation study on the parameters $\alpha/r$ to evaluate the sensitivity of the weighting of the LoRA in the balance between the original and style rewards within the reward model. Figure~\ref{fig:weights} clearly illustrates the trade-off between the environment reward and the style reward, where a higher weight results in a lower environment reward but an increase in style adaptation. The ablation study shows that FLoRA provides greater control over the balance between the original task and stylistic adaptation by adjusting a single weighting parameter, $\alpha$.

\subsubsection{EPIC Distance}
To measure the distance to the original reward function after the adaptation process in Section IV of the original paper, we utilize the Equivalent-Policy Invariant Comparison (EPIC) distance~\cite{gleave2020quantifying}. The results in Table~\ref{tab:epic} show that \textbf{FLoRA} retains the original reward at a greater rate while adapting to the new style in comparison to other baselines.

\begin{table}[t]
\centering
\caption{
    \textbf{EPIC Distance.} Comparison of the EPIC distance between the finetuned reward functions and the original reward function in the \texttt{Drawer-Close}. Lower values indicate closer alignment.}
\label{tab:epic}
\resizebox{1.0\columnwidth}{!}{
\begin{tabular}{lcccccc}
    \toprule
    \textbf{Method} & \textbf{FLoRA} & \textbf{SURF} & \textbf{Co-training} & \textbf{Fine-tuning} \\
    \midrule
    \rowcolor{white}
    \textbf{EPIC Distance} & \textbf{0.216} & 0.257 & 0.311 & 0.327 \\
    \bottomrule
\end{tabular}}
\end{table}

\begin{table*}[t]
    \centering
    \caption{Environment and Adapted Rewards of Environments from DeepMind Control Suite and Meta-World.}
    \label{tab:environments_rewards}
    \vspace{4pt}
     \renewcommand{\arraystretch}{1.25} % Adjust row spacing
    \setlength\arrayrulewidth{0.05pt} % Adjust width of dashed lines
    \resizebox{\textwidth}{!}{
    \begin{tabular}{l c c p{0.4\textwidth}}
        \toprule
        \textbf{Environment Source} & \textbf{Environment} & \textbf{Environment Reward} & \textbf{Adapted Reward} \tabularnewline
        \midrule
        \multirow{3}{*}{\textbf{DeepMind Control Suite} \cite{tunyasuvunakool2020dm_control}} 
        & \texttt{Walker} & Achieve stable, efficient bipedal locomotion to walk forward. & Linear reward based on the z-axis of the body. \tabularnewline 
        & \texttt{Cheetah} & Maximize speed along a track without falling over. & Linear reward based on the z-axis of the head. \tabularnewline 
        & \texttt{Quadruped} & Navigate varied terrains using four-legged locomotion. & Linear reward based on the z-axis of the body. 
        \vspace{10pt}
        \tabularnewline
        \arrayrulecolor{gray!20}\hdashline\arrayrulecolor{black}
        \tabularnewline
        \multirow{3}{*}{\textbf{Meta-World} \cite{yu2020meta}} 
        & \texttt{Button-Press} & Manipulate a robotic arm to press a button. & Negative reward as the robot arm is on the left side of the environment (min(x-axis position, 0)). \tabularnewline 
        & \texttt{Button-Press-Wall} & Press a button with a wall obstructing direct access. & Negative reward as the robot arm is on the left side of the environment (min(x-axis position, 0)). Navigate around a wall. \tabularnewline 
        & \texttt{Drawer-Close} & Manipulate a robotic arm to close a drawer. & Negative reward as the robot arm is on the left side of the environment (min(x-axis position, 0)). \tabularnewline
        \bottomrule
    \end{tabular}}
\end{table*}

\subsection{Implementation Details}
\label{appendix:implementation_details}

\subsubsection{Algorithm} We present the pseudo-code of FLoRA in Algorithm~\ref{alg:FLoRA}. For consistency, we start by training a policy $\policy_\omega$ from scratch for all experiments, but this is not a requirement: FLoRA can also be naturally employed with a pre-trained policy. In addition, we start from a pre-trained reward function $\hat{\reward}_{\psi_0}$ and a base RL algorithm $\mathcal{H}$. We employ SAC as the base RL algorithm, following \cite{lee2021pebble}. In Appendix~\ref{sec:hyperparameters} we present the hyperparameters used to define the LoRA. Empirically, $\alpha$ is set to match the initial value of $r$ and is not further adjusted. At each timestep we use the policy $\policy_\omega$ to collect experience tuples from the environment $(s_t, a_t, s_{t+1}, \hat{\reward}_{\psi_0,\psi_A,\psi_B}(s_t,a_t))$ (Line 3), where the reward value is computed using the forward pass of the adapted reward model as described in Equation~\ref{eq:lora}.  Every $M$ steps, we sample pairs of segments to collect human preferences in the form of feedback triples ($\sigma_0, \sigma_1, \zeta$) (Line 8-9) and update the LoRA matrices $\psi_A, \psi_B$ (Line 10).

\begin{algorithm}[t]
\caption{FLoRA: Preference-based RL via Low-Rank Style Adaptation} \label{alg:FLoRA}
\begin{algorithmic}[1]
    \REQUIRE Policy $\policy_\omega$, pre-trained reward function $\hat{\reward}_{\psi_0}$, base RL algorithm $\mathcal{H}$, learning rate $\alpha$, adaptation rank $r$, policy training interval $K$, human preference feedback interval $M$, total algorithm steps $N$
    \ENSURE Adapted reward function $\hat{\reward}_{\psi_0,\psi_A,\psi_B}$ and policy $\policy_\omega$
    
    \STATE \textbf{Initialize} adaptation parameters $\psi_A \in \mathbb{R}^{r \times k}$, where $\psi_{A_{ij}} \sim \mathcal{N}(0, \sigma^2)$, and $\psi_B = \mathbf{0}^{d \times r}$
    
    \FOR{$t = 1$ \textbf{to} $N$} 
        \STATE Collect experience buffer using policy $\policy_\omega$, ${\mathcal{B} \leftarrow \{(s_t, a_t, s_{t+1}, \hat{\reward}_{\psi_0,\psi_A,\psi_B}(s_t, a_t))\}_{j=1}^{B}}$
        
        \IF{$t \mod K = 0$}
            \STATE Update policy $\policy_\omega$ using base RL algorithm $\mathcal{H}$ with experience $\mathcal{B}$
        \ENDIF
        
        \IF{$t \mod M = 0$}
            \STATE Sample policy $\policy_\omega$ and form segment pairs $\{(\segment_0, \segment_1)\}_{i=1}^M$
            \STATE Collect human preferences over the segment pairs ${\dataset_{l_\text{new}} \leftarrow \dataset_{l_\text{new}} \cup \{(\segment_0, \segment_1, \zeta)\}_{i=1}^M}$
            \STATE Update adaptation parameters $\psi_A$ and $\psi_B$, following Equation~\ref{eq:cross},
        \ENDIF
        
    \ENDFOR
\end{algorithmic}
\end{algorithm}

\subsubsection{Robotic Tasks}

\texttt{Spot-Follow} is a motion task created for the four-legged Boston Dynamics' Spot to follow a human participant around a 3m$\times$3m room. To validate the policy as accurately as possible, we track both the robot and the human using MOCAP (motion capture), ensuring accurate readings so that any discrepancies primarily stem from the policy quality rather than from the perception pipeline. Spot was initially trained to follow a human from any direction in a simulation created in Unity. It is then fine-tuned with new preferences for style adaptation using FLoRA to follow from the positive x-axis. The policy was trained for 1 million timesteps, and preferences were collected in a similar fashion to the simulated experiments, where 10 preferences were given every 20K timesteps until a total of 100 was collected. The experiment measured how well Spot could move to the goal position when approaching the human. In this case, the participant stood in predetermined positions while Spot tried to reach the goal position relative to the participant's location. We transferred the final policy in a one-shot manner to the real robot. In total we collected trajectories from 20 different positions, making up for a total of 20 episodes per condition. Success rate (\%) is determined by whether Spot managed to get within a 0.5m radius of the goal position. Near collision rate (\%) is determined by whether it was necessary for Spot to activate its built-in collision avoidance to avoid walking into the human. Average finish distance ($m$) is measured as the closest distance to the goal point.

\texttt{RW-Close-Drawer} is a manipulation task created for the Franka robot to close a drawer. The setup is similar to Meta-Worlds'~\cite{yu2020meta} \texttt{Drawer-Close}, where the drawer is set up in front of Franka and variations come from moving the drawer sideways for each episode. The baseline reward model was trained using \texttt{Drawer-Close} with the task of closing the drawer. For the fine-tuned style behavior, it was further trained following FLoRA in \texttt{Drawer-Close} for a million timesteps to prefer a slower speed while closing. We collected 10 preferences per 10K timesteps until we reached a total of 200 preferences. Similarly to \texttt{Spot-Follow}, we transferred the final policy in a one-shot manner to the real robot, and in total, we collected 6 episodes per condition to evaluate the metrics. Distance Pushed ($cm$) measures how far Franka pushed in the drawer on average. Velocity ($ms^{-1}$) measures the average velocity across the whole episode. 

\subsubsection{Baselines} For policy learning across all methods, we use PEBBLE~\cite{lee2021pebble}, a state-of-the-art preference-based RL algorithm. We update the reward function every $M = 20$K step, with each feedback update utilizing $10\%$ of the total queries until all queries are acquired. We follow the standard practice~\cite{christiano2017deep,lee2021pebble} and model the standard reward function as an MLP binary classifier. Queries are sampled by ensemble disagreement similarly to \cite{hejna2023few,lee2021pebble}. FLoRA fine-tunes low-rank matrices $\psi_A$ and $\psi_B$ for each frozen hidden layer in the original reward model. All the hyperparameters used for both the policies and reward functions can be consulted in Appendix~\ref{sec:hyperparameters}.

\nopar{\textbf{Fine-tuning}: As the default baseline, we employ simple fine-tuning of the complete reward model to novel human preferences, an approach akin to~\cite{hejna2023few} for meta-learning.}

\nopar{\textbf{SURF}~\cite{park2022surf}: A semi-supervised learning approach that employs data augmentation to improve reward function generalization and mitigate CRF. SURF generates preferences through \emph{pseudo-labeling}, labeling unlabeled queries that exhibit a high degree of confidence $\tau$ according to a reward model ensemble, and \emph{temporal cropping}, creating new queries with reduced trajectory lengths while preserving the preference from labeled queries. We employ the code provided by the authors as well as the suggested hyperparameters in~\cite{park2022surf}.}

\nopar{\textbf{Co-training}: In this baseline we re-use the previously acquired dataset of the original reward and the new style samples to fine-tune the model, similar to DAGGER~\cite{ross2011reduction}.}

\subsubsection{Human Adaptation Styles} 
\label{sec:styles}
In each environment, the reward function must adapt to a new objective. The reward utilized by the synthetic oracles is calculated as the sum of both components. Synthetic human preferences are determined based on the cumulative sum of the reward in relation to the environmental reward, featuring an error rate of $\sim10\%$. To obtain $\psi_0$ we pre-train $\reward_\psi$ with 500 queries for the DeepMind Control Suite environments, 20K for Button Press and Button Press Wall, and 5K for Drawer Close from Meta-World.

\subsubsection{Hyperparameters}
\label{sec:hyperparameters}

In Table~\ref{tab:policy-hyperparameters} we present the hyperparameters used for PEEBLE and SURF in regards to the RL training and network architecture of the agents. In Table~\ref{tab:reward-hyperparameters} we present the hyperparameters and network architecture of the reward functions employed in this work.

\begin{table*}[t]
\centering
\caption{Hyperparameters for PEBBLE and SURF in Different Environments}
\centering
\label{tab:policy-hyperparameters}
\resizebox{\textwidth}{!}{
\begin{tabular}{lcccc}
    \toprule
    & \textbf{DM Control - PEBBLE} & \textbf{Meta-World - PEBBLE} & \textbf{DM Control - SURF} & \textbf{Meta-World - SURF} \\
    \midrule
    \rowcolor{gray!20} \multicolumn{5}{c}{\textbf{Network Architecture}} \\
    \midrule
    \{Hidden, Output\} Activation (Actor) & \{ReLU, Tanh\} & \{ReLU, Tanh\} & \{ReLU, Tanh\} & \{ReLU, Tanh\} \\
    \{Hidden, Output\} Activation (Critic) & \{ReLU, Tanh\} & \{ReLU, Tanh\} & \{ReLU, Tanh\} & \{ReLU, Tanh\} \\
    Hidden Layers (Actor) & \{1024, 1024\} & \{256, 256, 256\} & \{256, 256, 256\} & \{256, 256, 256\} \\
    Hidden Layers (Critic) & \{1024, 1024\} & \{256, 256, 256\} & \{256, 256, 256\} & \{256, 256, 256\} \\
    
    \midrule
    \rowcolor{gray!20} \multicolumn{5}{c}{\textbf{Reinforcement Learning Parameters}} \\
    \midrule
    Critic Learning Rate & 5e-4 & 5e-4 & 5e-4 & 5e-4 \\
    Actor Learning Rate & 5e-4 & 5e-4 & 5e-4 & 5e-4 \\
    Batch Size & 256 & 512 & 256 & 512 \\
    Discount Factor & 0.99 & 0.99 & 0.99 & 0.99 \\
    Learning Starts & 9000 & 9000 & 9000 & 9000 \\
    Tau & 0.005 & 0.005 & 0.005 & 0.005 \\
    Gamma & [0.9, 0.999] & [0.9, 0.999] & [0.9, 0.999] & [0.9, 0.999] \\
    Actor Update Frequency & 1 & 1 & 1 & 1 \\
    Critic Update Frequency & 2 & 2 & 2 & 2 \\
    Gradient Steps & 1 & 1 & 1 & 1 \\
    
    \bottomrule
\end{tabular}
}
\end{table*}

\begin{table}[t]
\centering
\caption{Shared Hyperparameters Across Reward Functions}
\label{tab:reward-hyperparameters}
\resizebox{0.45\textwidth}{!}{
\begin{tabular}{lc}
    \toprule
    \rowcolor{gray!20} \multicolumn{2}{c}{\textbf{Network Architecture}} \\
    \midrule
    \{Hidden, Output\} Activation & \{ReLU, Tanh\} \\
    Hidden Sizes & \{256, 256, 256\} \\
    
    \midrule
    \rowcolor{gray!20} \multicolumn{2}{c}{\textbf{Preference Learning}} \\
    \midrule
    Segment Size & 50 \\
    Initial Amount of Queries & 1/10 of queries (1/80 Meta-World) \\
    Initial Training Epochs & 200 \\
    Queries per Update & 1/10 of queries (1/80 Meta-World) \\
    Training Epochs per Update & 50 (10 Meta-World) \\
    Batch Size & 128 \\
    Timesteps Between Updates & 20K (5K Meta-World) \\
    Learning Rate & 0.0003 \\
    
    \midrule
    \rowcolor{gray!20} \multicolumn{2}{c}{\textbf{FLoRA}} \\
    \midrule
    Rank & 16 \\
    Lora Weight & 16 \\
    
    \midrule
    \rowcolor{gray!20} \multicolumn{2}{c}{\textbf{SURF Hyperparameters}} \\
    \midrule
    Segment Size & 60 \\
    Unlabeled Batch Ratio & 4 \\
    Threshold & 0.99 \\
    Loss Weight & 1 \\
    Crop Length & 50 \\
    
    \bottomrule
\end{tabular}}
\end{table}

\subsubsection{Code, Videos and Hardware Utilized} Our FLoRA implementation is available \href{https://github.com/DanielLSM/flora}{here}. We include the weights for a pre-trained reward function tailored for Walker, along with detailed instructions for executing the code. Several videos showcasing policies across some environments and conditions described in the main paper can be viewed \href{https://sites.google.com/view/preflora/}{here}. All baseline experiments and ablations were conducted on a workstation equipped with a 16-core CPU Ryzen 5950x, a 12GB NVIDIA RTX 3080 with 8960 CUDA cores, and 64GB of DDR4 RAM. All SURF and SURF + FLoRA experiments and ablations were conducted on server instances using Azure. Each experiment was run on a container instance featuring 1 vCPU and 12GB RAM.

\end{document}

%% file: settings.tex
% # Preference learning #
\newcommand{\trajectory}{\tau}
\newcommand{\highlight}{h}
\newcommand{\highlightdataset}{\mathcal{D}}
\newcommand{\highlightdiscount}{\lambda}
\newcommand{\highlightweight}{\alpha} 
\newcommand{\highlightlength}{L}
\newcommand{\segment}{\sigma}
\newcommand{\ntrajectories}{l}
\newcommand{\segmentdataset}{\mathcal{D}_{\sigma}}
\newcommand{\nsegments}{k}
\newcommand{\rewardnetwork}{\hat{\reward}_{\psi}}
\newcommand{\rewardnetworkobservation}{\rewardnetwork(s,a)}
\newcommand{\loss}{\mathcal{L}}
\newcommand{\dataset}{\mathcal{D}}
\newcommand{\preference}{\zeta}
\newcommand{\query}{(\segment^0,\segment^1,\zeta)}
\newcommand{\datasettraj}{\mathcal{D_{\trajectory}}}
\newcommand{\datasetquery}{\mathcal{D}_q}
\newcommand{\preferencedataset}{\mathcal{D}}
\newcommand{\highlighteddataset}{\mathcal{D_{hq}}}

% # Active Learning #
\newcommand{\stateseq}{\bm{s}_{\segment}}

% # RL #
\newcommand{\reward}{r}
\newcommand{\policy}{\pi}
\newcommand{\States}{\mathcal{S}}
\newcommand{\s}{s}
\newcommand{\Actions}{\mathcal{A}}
\newcommand{\action}{a}
\newcommand{\safeActions}{{\Actions_\mathrm{safe}}}

% # MATH #
\newcommand{\naturalnumb}{\mathbb{N}_+}
\newcommand{\realnumb}{\mathbb{R}}

% # Abreviations #
\newcommand{\ie}{i.e., }
\newcommand{\eg}{e.g., }

% # Language stuff #
\newcommand{\feature}{f}
\newcommand{\prompt}{\text{prompt}}
\newcommand{\chatgpt}{\text{LLM}}
\newcommand{\response}{\text{r}_i}
\newcommand{\sentiment}{\text{y}}
\newcommand{\val}{\text{v}}
\newcommand{\mapCarnationPinkping}{\text{M}}
\newcommand{\tensor}{\mathcal{T}}
\newcommand{\metric}{\text{m}}